\documentclass[1p]{elsarticle}

\usepackage{algorithm}
\usepackage{algorithmic}
\usepackage{amsfonts}
\usepackage{booktabs}
\usepackage{amsmath}
\usepackage{subfigure}
\usepackage{multirow}
\usepackage [figuresright] {rotating}
\usepackage{appendix}
\newdefinition {problem}{Problem}
\newdefinition {definition} {Definition}

\begin{document}

\makeatletter
\def\ps@pprintTitle{%
   \let\@oddhead\@empty
   \let\@evenhead\@empty
   \let\@oddfoot\@empty
   \let\@evenfoot\@oddfoot
}
\makeatother

\begin{frontmatter}
\title{Towards Non-I.I.D. Image Classification: A Dataset and Baselines}

\author [1] {Yue He \fnref {fn1}}
\ead {heyue18@mails.tsinghua.edu.cn}

\author [1] {Zheyan Shen \fnref {fn1}}
\ead {shenzy17@mails.tsinghua.edu.cn}

\author [1] {Peng Cui \corref{cor1} \fnref {fn2}}
\ead {cuip@tsinghua.edu.cn}

\cortext[cor1]{Corresponding author}

\fntext [fn1] {Ph.D candidate, Department of Computer Science and Technology, Tsinghua University}
\fntext [fn2] {Associate Professor (Tenured), Department of Computer Science and Technology, Tsinghua University}

\address [1]{Lab of Media and Network, Room 9-316, East Main Building, Tsinghua University, Beijing 100084, P.R.China}

\begin{abstract}
I.I.D.\footnote{\textbf{I.I.D.}: Independent and Identically Distributed} hypothesis between training and testing data is the basis of numerous image classification methods. Such property can hardly be guaranteed in practice where the Non-IIDness is common, causing instable performances of these models. In literature, however, the Non-I.I.D.\footnote{\textbf{Non-I.I.D}: Non-Independent and Identically Distributed} image classification problem is largely understudied. A key reason is lacking of a well-designed dataset to support related research.
In this paper, we construct and release a Non-I.I.D. image dataset called NICO\footnote{\textbf{NICO}: Non-I.I.D. Image dataset with Contexts}, which uses contexts to create Non-IIDness consciously. 
Compared to other datasets, extended analyses prove NICO can support various Non-I.I.D. situations with sufficient flexibility.
Meanwhile, we propose a baseline model with ConvNet structure for General Non-I.I.D. image classification, where distribution of testing data is unknown but different from training data. The experimental results demonstrate that NICO can well support the training of ConvNet model from scratch, and a batch balancing module can help ConvNets to perform better in Non-I.I.D. settings.

\end{abstract}
    

\end{frontmatter}

\section{Introduction}
In recent years, machine learning has achieved remarkable progress, mainly owing to the development of deep neural networks \cite{NIPS2012_4824, Simonyan2014Very, he2016deep, Ren2015Faster, long2015fully, ma2018progressive}. 
One basic hypothesis of machine learning models is that the training and testing data should consist samples Independent and Identically Distributed (I.I.D.). 
However, this ideal hypothesis is fragile in real cases where we can hardly impose constraints on the testing data distribution. 
This implies that the model minimizing empirical error on training data does not necessarily perform well on testing data, leading to the challenge of Non-I.I.D. learning. 
The problem is more serious when the training samples are not sufficient to approximate the training distribution itself. 
How to develop Non-I.I.D. learning methods that are robust to distribution shifting is of paramount significance for both academic research and industrial applications.

Benchmark datasets, providing a common ground for competing approaches, are always important to promote the development of a research direction. 
Take image classification, a prominent learning task, as an example. 
Its development benefits a lot from the benchmark datasets, such as PASCAL VOC \cite {Everingham2015The}, MSCOCO \cite {Lin2014Microsoft}, and ImageNet \cite {Deng2009ImageNet}. 
In particular, it is the ImageNet, a large-scale and well-structured image dataset, that successfully demonstrates the capability of deep learning and thereafter significantly accelerates the advancement of deep convolutional neural networks. 
On these datasets, it is easy to establish an I.I.D. image classification setting by random data splitting.  
But they do not provide an explicit option to simulate a Non-I.I.D. setting. 
The dataset that can well support the research on Non-I.I.D. image classification is still in vacancy.   

In this paper, we construct and release a dataset that is dedicately designed for Non-I.I.D. image classification, named NICO (Non-I.I.D. Image dataset with Contexts). 
The basic idea is to label images with both main concept and contexts. 
For example, in the category of `dog', images are divided into different contexts such as `grass', `car', `beach', meaning the `dog' is on the grass, in the car, or on the beach respectively. 
With these contexts, one can easily design an Non-I.I.D. setting by training a model in some contexts and testing it in the other unseen contexts. 
Meanwhile, the degree of distribution shift can be flexibly controlled by adjusting the proportions of different contexts in training and testing data. 
Till now, NICO contains 19 classes, 188 contexts and nearly 25,000 images in total. 
The scale is still increasing, and the current scale has been able to support the training of deep convolution networks from scratch.

The NICO dataset can support, but not limited to, two typical settings of Non-I.I.D. image classification. 
One is Targeted Non-I.I.D. image classification, where testing data distribution is known but different from training data distribution. 
The other is General Non-I.I.D. image classification, where testing data distribution is unknown and different from training data distribution. 
Apparently, the latter one is much more realistic and challenging. 
A model learned in one environment could be possibly applied in many other environments. 
In this case, the robustness of a model in the environments with unknown distribution shift is a highly favorable characteristic. 
It is especially critical in risk-sensitive applications like medical and security.

Due to the lack of a well-structured and reasonable-scaled dataset, there is still no convolutional neural network model proposed to address the general Non-I.I.D. image classification problem. 
In this paper, we propose a novel model CNBB\footnote{\textbf{CNBB}: ConvNet with Batch Balancing} (ConvNet with Batch Balancing) as a baseline of exploiting CNN model for general Non-I.I.D. image classification.The experimental results show that the proposed batch balancing mechanism can help a ConvNet model to resist, to some extent, the negative effect brought by Non-IIDness. 
\section{Non-I.I.D. Image Classification}

\subsection{Problem Definition}
We first give a formal definition of Non-I.I.D. image classification as follow:
\begin{problem}
\label{problem1}
\textbf{(Non-I.I.D. Image Classification)}
Given the training data $D_{train}=(X_{train},Y_{train})$, where $X_{train}\in\mathbb{R}^{n\times (c\times h\times w)}$ represents the images and $Y_{train}\in\mathbb{R}^{n\times 1}$ represents the labels.
The task is to learn a feature extractor $g_\varphi(\cdot)$ and a classifier $f_\theta(\cdot)$, so that $f_\theta(g_\varphi(\cdot))$ can predict the labels of testing data $D_{test}=(X_{test},Y_{test})$ precisely, where $g_\varphi(\cdot)\in\mathbb{R}^{n\times p}$ and $\psi(D_{train})\neq\psi(D_{test})$.
Moreover, according to the availability of the prior knowledge on testing data, we further define two different tasks. 
One is \textbf{Targeted Non-I.I.D. Image Classification} where the testing data distribution $\psi(D_{test})$ is known.
The other is \textbf{General Non-I.I.D. Image Classification}, which corresponds to a more realistic scenario where the testing data distribution $\psi(D_{test})$ is unknown.
\end{problem}

In order to intuitively quantify the degree of distribution shift between $\psi(D_{train})$ and $\psi(D_{test})$, we define the Non-I.I.D. Index as follow:
\begin{definition}
\label{definition1}
\textbf{Non-I.I.D. Index (NI)}
Given a feature extractor $g_\varphi(\cdot)$ and a class $C$, the degree of distribution shift between training data $D_{train}^C$ and testing data $D_{test}^C$ is defined as:
\begin {small}
$$NI(C)=\left\|\frac {\overline{g_\varphi(X_{train}^C)} - \overline{g_\varphi(X_{test}^C)}}{\sigma (g_\varphi(X^C))} \right\|_2,$$
where $X^C=X_{train}^C \cup X_{test}^C$, $\overline{(\cdot)}$ represents the first order moment, $\sigma (\cdot)$ is the std used to normalize the scale of features and $\left\|\cdot\right\|_2$represents the 2-norm.
\end {small}
\end{definition}

\subsection{Existence of Non-IIDness}

\begin{figure}[t]
\centering 
\includegraphics[width=.64\linewidth]{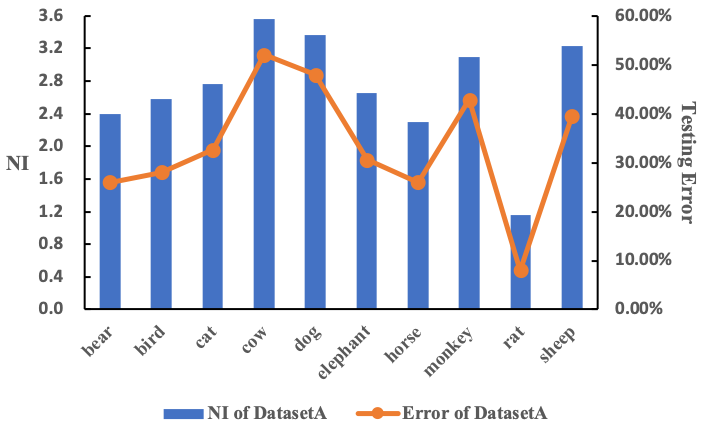}
\caption{$NI$ (represented by the bar-type) and testing error (represented by the curve-type) of each class in Dataset A.}
\label{fig:na}
\end{figure}

In real cases, the I.I.D. hypothesis can never be strictly satisfied, meaning that Non-IIDness ubiquitously exists in previous datasets \cite {torralba2011unbiased}. 
Here we take ImageNet as an example. 
ImageNet is in a hierarchical structure, where each class (e.g. dog) contains multiple subclasses (e.g. different kinds of dogs). 
For each subclass, it provides training and testing (validation) subsets of images.
To verify the Non-IIDness in ImageNet, we select 10 common animal classes (e.g. dog, cat) and construct a new dataset using 10 instantiated subclasses (e.g. 
Labrador, Persian), each randomly drawn from those classes. 
Using the training and testing subsets, we train and evaluate a ConvNet on image classification task.
The structure of the ConvNet used in this paper is similar to AlexNet (details seen in \textbf{Appendix}), and we take the last FC layer of the ConvNet as the feature extractor $g_\varphi$.
Note that model structure is used in all subsequent analysis (including on NICO) for fair comparison, and thus selected by trading-off performance and required training data scale. 
But as a base model with sufficient learning capacity, the specific model structure does not affect the conclusions.
We repeat this collection procedure for 3 times, obtain 3 new datasets ($Dataset\ A$, $Dataset\ B$ and $Dataset\ C$) and calculate the $NI$ and testing error for each class respectively.
As an example, we plot the results of $Dataset A$ in Figure \ref{fig:na}.
We can find that:

\begin{itemize}
    \item $NI$ is above zero for all classes, which implies the Non-IIDness between training and testing data is ubiquitous even in large-scale datasets like ImageNet.
    \item Different classes have different $NI$ values and higher $NI$ value corresponds to higher testing error.
\end{itemize}

The strong correlation between $NI$ and testing error can be further proved by their high pearson correlation \cite{tutorials2014pearson} coefficients ($r=0.95$) and small $p\_value$ (2e-15).
The showcase and statistical analysis well support an plausible conclusion that the degree of distribution shift quantified by $NI$ is a key factor influencing classification performance.
Although the numerical value of $NI$ is conditioned on specific feature extractor, we could use it to analyse the trend of distribution bias by some intervention between training and testing data, if feature extractor is fixed.
In later paragraph, we use $NI$ to make an empirical analysis on the new dataset we construct to prove that NICO can support various Non-I.I.D. situations flexibly and consciously.

\begin{figure}[t]
\centering  
\includegraphics[width=.64\linewidth]{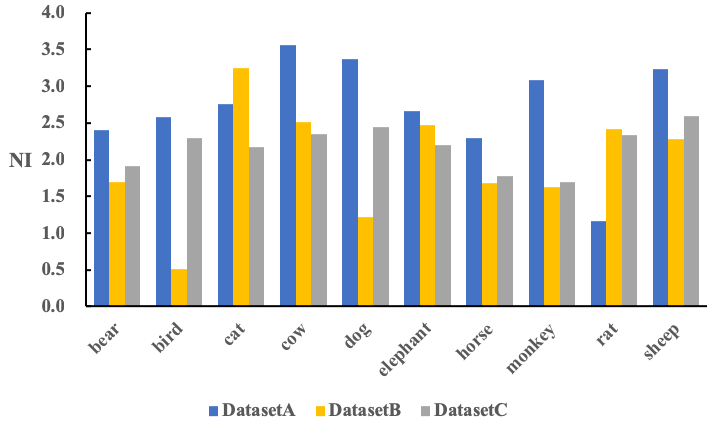}
\caption{$NI$ of each class in 3 different datasets constructed from ImageNet. Different datasets instantiate the same classes with different subclasses.}
\label{fig:ni}
\end{figure}

\subsection{Limitations of Existing Datasets}
Throughout the development of computer vision research, benchmark datasets have always played a critical role on both providing a common ground for algorithm evaluation and driving new directions.
Specifically, for image classification task, we can enumerate several milestone datasets such as PASCAL VOC, MSCOCO and ImageNet.
However, existing benchmark datasets cannot well support the Non-I.I.D image classification.
First of all, despite the manifested Non-IIDness in ImageNet and other datasets, as shown in Figure \ref{fig:na}, the overall degree of distribution shift between training and testing data for each class is relatively small, making these datasets less challenging from the angle of Non-I.I.D. image classification. 
More importantly, there is no explicit way to control the degree of distribution shift between training and testing data in the existing datasets.
As illustrated in Figure \ref{fig:ni}, if we instantiate the same class with different subclasses in ImageNet and obtain 3 datasets with identical structure, the $NI$ of a given class is fairly unstable across different datasets.
Without a controllable way to simulate different levels of Non-IIDness, competing approaches cannot be evaluated fairly and systematically on those datasets. 
Those said, a dataset that is dedicatedly designed for Non-I.I.D. image classification is demanded.

\section{The NICO Dataset}
In this section, we introduce the properties and collection process of the dataset, followed by preliminary empirical results in different Non-I.I.D. settings supported by this dataset. 

\subsection{Context for Non-I.I.D. Images}
The essential idea of generating Non-I.I.D. images is to enrich the labels of an image with both conceptual and contextual labels. 
Different from previous datasets that only label an image with the major concept (e.g. dog), we also label the concrete context (e.g. on grass) that the concept appears in. 
Then it is easy to simulate an Non-I.I.D. setting by training and testing the model of a concept with different contexts. 
A good model for Non-I.I.D. image classification is expected to perform well in both training contexts and testing contexts.

\begin{figure*}[t]
\centering 
\hspace*{-0.1\linewidth}
\includegraphics[width=1.2\linewidth]{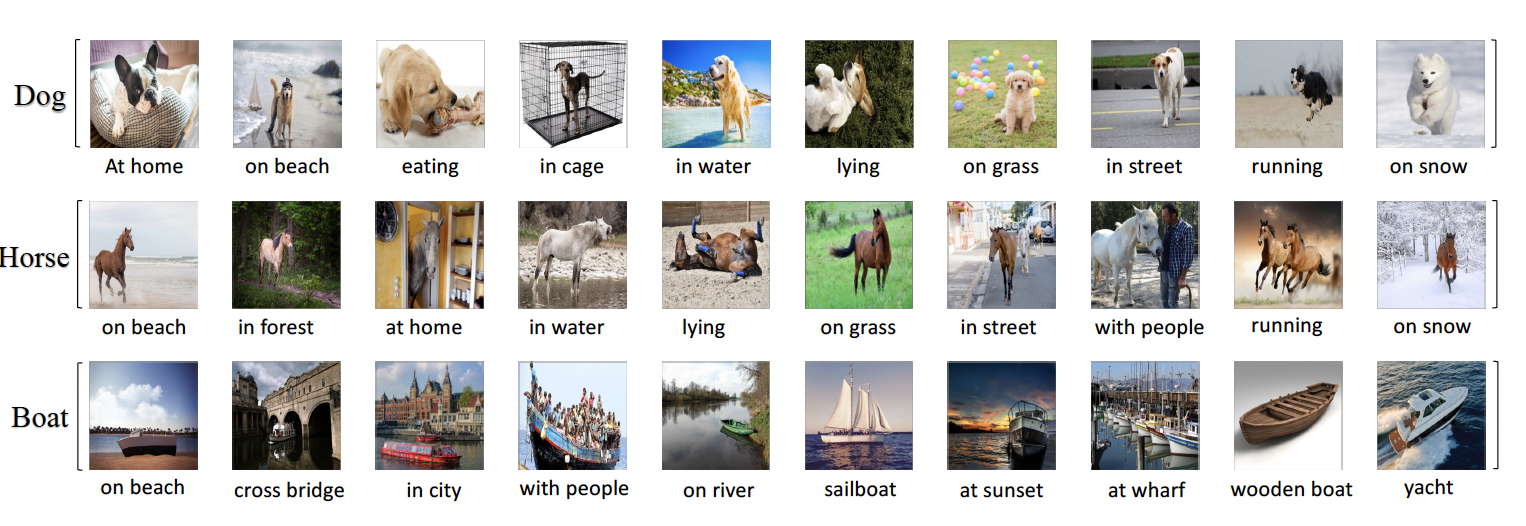}
\caption{Samples with contexts in NICO. Images in the first row are dogs of $Animal$, assigned to different contexts below it. The second and third row correspond to horse of $Animal$ and boat of $Vehicle$ respectively.}
\label{fig:samples}
\end{figure*}

In NICO, we mainly incorporate two kinds of contexts. 
One is the attributes of a concept (or object), such as color, action, and shape. 
Some examples of `context + concept' pairs include \textit{white bear}, \textit{climbing monkey} and \textit{double decker} etc. 
The other kind of contexts is the background or scene of a concept. 
The examples of `context + concept' pairs include \textit{cat on snow}, \textit{horse aside people} and \textit{airplane in sunrise} etc. 
Samples of different contexts in the NICO dataset are shown in Figure \ref{fig:samples}.

In order to provide more flexible Non-I.I.D. settings, we tend to select the contexts that occur in multiple concepts. 
Then for a given concept, a context may occur in both positive samples and negative samples (that are sampled from other concepts). 
This provides another flexibility to let a context included in training positive samples appear or do not appear in training negative samples, which will yield different Non-I.I.D. settings.

\subsection{Data Collection and Statistics}
Referring to ImageNet, MSCOCO and other classical datasets \cite{krizhevsky2009learning, Kuznetsova2018The}, we first confirm two superclasses: $Animal$ and $Vehicle$.
For each superclass, we select classes from the 272 candidates in MSCOCO, with the criterion that the selected classes in a superclass should have large inter-class differences.
For context selection, we exploit YFCC100m\cite{thomee2015yfcc100m} broswer\footnote{http://www.yfcc100m.org/} and first derive the frequently co-occurred tag list for a given concept (i.e. class label). 
We then filter out the tags that occur in only a few concepts. 
Finally, we manually screen all tags and select the ones that are consistent with our definition of contexts (i.e. object attributes or backgrounds and scenes). 

\begin{table}[htbp]
\caption{Data size of each class in NICO.}
\label{table1}
\begin{center}
\begin{small}
\begin{sc}
\begin{tabular}{lc|lc}
\toprule
$Animal$ & Data Size & $Vehicle$ & Data Size \\
\midrule
Bear & 1609 & Airplane & 930 \\
Bird & 1590 & Bicycle & 1639 \\
Cat & 1479 & Boat & 2156 \\
Cow & 1192 &Bus & 1009 \\
Dog & 1624 & Car& 1026 \\
Elephant & 1178 & Helicopter & 1351 \\
Horse & 1258 & Motorcycle & 1542 \\
Monkey & 1117 & Train & 750 \\
Rat & 846 &Truck & 1000 \\
Sheep & 918 &  &  \\
\bottomrule
\end{tabular}
\end{sc}
\end{small}
\end{center}
\end{table}

After obtaining the conceptual and contextual tags, we concatenate a given conceptual tag and each of its contextual tags to form a query, input the query into the API of Google and Bing image search, and collect the top-ranked images as candidates. 
Finally, in the phase of screening, we select images into the final dataset according to the following criteria: 

\begin{itemize}
    \item The content of an image should correctly reflects its concept and context.
    \item Given a class, the number of images in each context should be adequate and as balance as possible across contexts. 
\end{itemize}

Note that we do not conduct image registration or filtering by object centralization, so that the selected images are more realistic and in wild than those in ImageNet. 

The NICO dataset will be continuously updated and expanded. 
Till now, there are two superclasses: $Animal$ and $Vehicle$, with 10 classes for $Animal$ and 9 classes for $vehicle$.
Each class has 9 or 10 contexts. 
The average size of contexts per class ranges from 83 to 215, and the average size of classes is about 1300 images, which is similar to ImageNet. 
In total, there are 25,000 images in the NICO dataset. 
As NICO is in a hierarchical structure, it is easy to be expanded.
More statistics on NICO is reported in Table \ref{table1}. 
The dataset can be downloaded through the link\footnote{https://www.dropbox.com/sh/8mouawi5guaupyb/AAD4fdySrA6fn3PgSmhKwFgva?dl=0} or the link\footnote {https://pan.baidu.com/s/1277mgM-Nju6REd5h3xXlrA} for Chinese.

\subsection{Supported Non-I.I.D. Settings}
By dividing a class into different contexts, NICO provides the flexibility of simulating Non-I.I.D. settings in different levels. 
To name a few, here we list 4 typical settings.

\begin{enumerate} [\textbf{Setting 1.}]
    \item [\textbf{Setting 1.}] \textbf{Minimum bias}. Given a class, we can ignore the contexts, and randomly split all images of the class into training and testing subsets as positive samples. Then we can randomly sample images belonging to other classes into training and testing subsets as negative samples. In this setting, the way of random sampling lead to minimum distribution shift between training and testing distributions in the dataset, which simulates a nearly i.i.d. scenario.
    \item [\textbf{Setting 2.}] \textbf{Proportional bias}. Given a class, when sampling positive samples, we use all contexts for both training and testing, but the percentage of each context is different in training and testing subsets. For example, we can let one context take the majority in training data while taking minority in testing, which  is consistent with the natural phenomena that visual concepts follow a power law distribution\cite{clauset2009power}.The negative sampling process is the same as Setting 1. In this setting, the level of distribution shift can be tuned by adjusting the proportion difference between training and testing subsets for each context.
    \item [\textbf{Setting 3.}] \textbf{Compositional bias}. Given a class, not every testing context that the positive samples belong to appears in training subset simultaneously.Such a setting is quite common in real scene, because available datasets could not contain all the potential contexts in nature due to the limitations of sampling time and space.Intuitively, the distribution shift from observed contexts to unseen contexts is usually large. The less number of testing contexts observed in training generally leads to the higher distribution shift.A more radical distribution shift can be further achieved by combining compositional bias and proportional bias.
    \item [\textbf{Setting 4.}] \textbf{Adversarial bias}. Given a class, the positive sampling process is the same as Setting 3. For negative sampling, we tend to select the negative samples from the contexts that have not been (or have been) included in positive training samples to form the negative training (or testing) subset. In this way, the distribution shifting is even higher than Setting 3, and the existing classification model developed under i.i.d. assumption are more prone to be confused.
\end{enumerate}

The above 4 settings are designed to generate Non-I.I.D. training and testing subsets. 
Under each setting, we can conduct either Targeted or General Non-I.I.D. image classification by assuming the distribution of testing subset is known or unknown.

\subsection{Empirical Analysis}
To verify the effectiveness of NICO in supporting Non-I.I.D image classification, we conduct a series of empirical analysis.
It is worth noting that, in each setting, only the distribution of training or testing data change, while the structure of ConvNet and the size of training data keep the same.

\begin{figure}[t]
\centering
\subfigure[Average $NI$ over all classes in $Animal$ superclass with respect to various dominant ratio of training data, while the dominant ratio of testing data is fixed to 1:1 (uniform sampling).]{
\begin{minipage}{1.\linewidth}
\centering
\includegraphics[width=.6\linewidth]{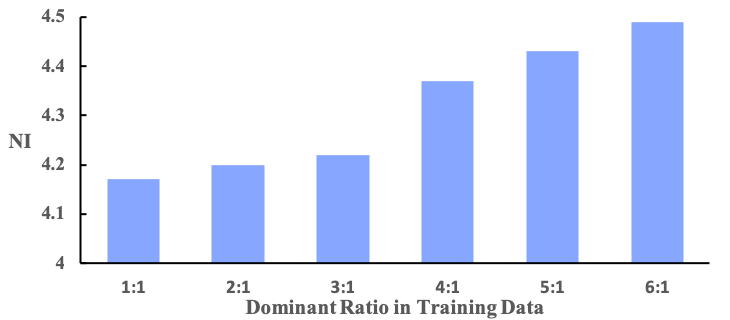}
%\caption{fig1}
\end{minipage}%
}%

\subfigure[Average $NI$ over all classes in $Animal$ superclass with respect to various dominant ratio of testing data, while the dominant ratio of training data is fixed to 5:1.]{
\begin{minipage}{1.\linewidth}
\centering
\includegraphics[width=.6\linewidth]{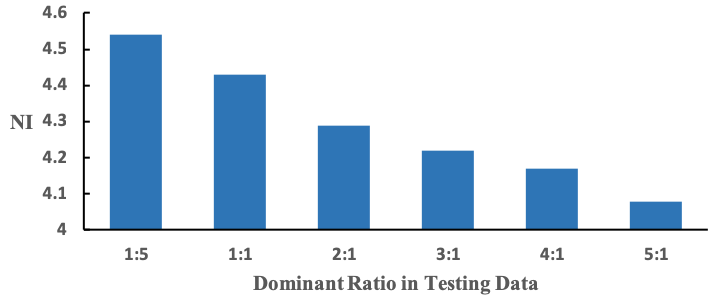}
%\caption{fig1}
\end{minipage}%
}%
\centering
\caption{$NI$ in proportional bias setting.}
\label {fig:a2}
\end{figure}

\subsubsection{Minimum Bias Setting}
In this setting, we randomly sample 8000 images for training and 2000 images for testing from $Animal$ and $Vehicle$ superclasses respectively.
The average testing accuracy and $NI$ over all the classes are $49.6\%$, $3.85$ for $Animal$ superclass and $63.0\%$, $3.20$ for $Vehicle$ superclass.
We can find that $NI$ in NICO is much higher than $NI$ in ImageNet even if there is no explicit bias (due to random sampling) when we construct the training and testing subsets.
This is because the images in NICO are typically non-iconic images with rich contextual information and non-canonical viewpoints, which is more challenging from the perspective of image classification.

% \begin{table}[h]
% \caption{Average testing accuracy and $NI$ of superclasses Animal and Vehicle under minimun bias setting.}
% \label{table2}
% \begin{center} 
% \vskip 0.1in
% \setlength{\tabcolsep}{6mm}{
% \begin{tabular}{|l|c|c|}
% \hline
%  & Testing Accuracy & $NI$ \\
% \hline 
% Animal & 49.60\% & 3.85 \\
% \hline
% Vehicle & 63.00\% & 3.20 \\
% \hline
% \end{tabular}}
% \end{center}
% \end{table}

\begin{figure}[t]
\centering 
\includegraphics[width=.6\linewidth]{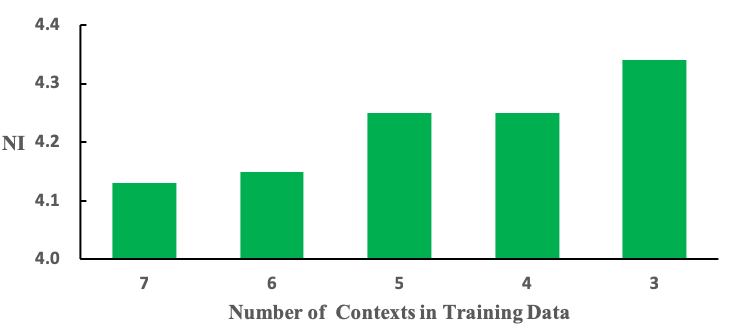}
\caption{$NI$ in compositional bias setting: average $NI$ over all classes in $Vehicle$ superclass with respect to the number of contexts used in training data.}
\label{fig:v3_1}
\end{figure}   

\begin{figure}[htbp]
\centering 
\includegraphics[width=.6\linewidth]{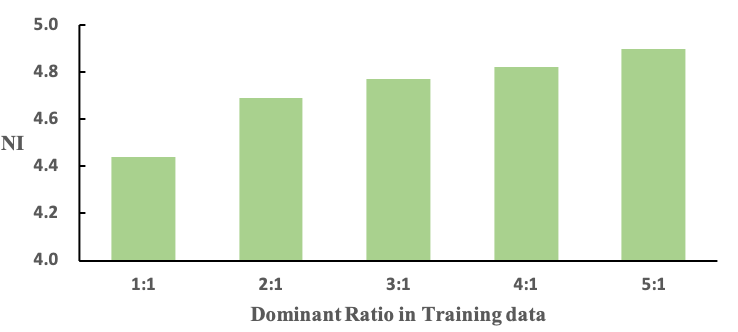}
\caption{$NI$ in the combined setting of compotisional bias and proportional bias: average $NI$ over all classes in $Vehicle$ superclass with respect to various dominant ratio of training data, where contexts in testing data is totally unseen in training.}
\label{fig:v3_2}
\end{figure}

\subsubsection{Proportional Bias Setting}
In this setting, we let all the contexts appear in both training and testing data, and randomly select one dominant context in training data (or testing data) for each class in $Animal$ superclass.
Such experimental settings comply with the natural phenomena that a majority of visual contexts are rare except a few common ones \cite {clauset2009power}.
Specifically, we define the dominant ratio as follow:
$$Dominant\ Ratio = \frac{N_{dominant}}{N_{minor}},$$
where $N_{dominant}$ refers to the sample size of the dominant context and $N_{minor}$ refers to the average size of other contexts where we uniformly sample other contexts.
We conduct two experiments where either dominant ratio of training data or testing data is fixed, and vary the other one.
We plot the results in Figure \ref{fig:a2} (a) and Figure \ref{fig:a2} (b).
From the figures, we can clearly find a consistent pattern that the $NI$ becomes higher as the discrepancy between dominant ratio of training data and testing data becomes larger.
As a result, by tuning the dominant ratio of training data (or testing data), we can easily simulate different extents of distribution shift as we want.

\begin{figure} [htbp]
\centering 
\includegraphics[width=.64\linewidth]{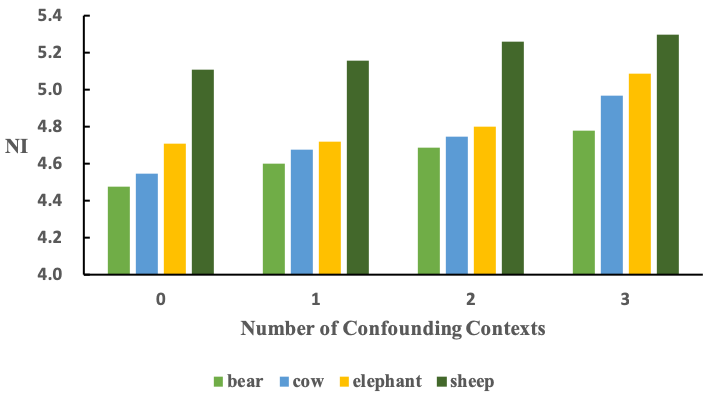}
\caption{NI in the adversarial bias setting: $NI$ of target class with respect to the number of confounding contexts.}
\label{fig:a4}
\end{figure}

\subsubsection{Compositional Bias Setting}
Compared to proportional bias setting, compositional bias setting simulates a condition where the knowledge obtained from training data is insufficient to characterize the whole distribution.
To doing so, we choose a subset of contexts for a given class when constructing the training data and test the model with all the contexts.
By varying the number of contexts observed in training data, we can simulate different extents of information loss and distribution shift.
From Figure \ref{fig:v3_1}, we can find that the $NI$ consistently decreases when we observed more contexts in training data.
A more radical distribution shift can be achieved by combining the notion of proportional bias and compositional bias.
Given a particular class in $Vehicle$ superclass, We choose 7 contexts for training and the other 3 contexts for testing, and further let one context dominate the training data.
By doing so, we can obtain a more severe Non-I.I.D. condition between training and testing data than previous two settings, as illustrated by the results from Figure \ref{fig:v3_2}.

\subsubsection{Adversarial Bias Setting}
Given a target class, we define a context as confounding context if it only appears in the negative samples of training data and positive samples of testing data.
In this experiment, we choose four classes in $Animal$ superclass as target classes and report the $NI$ w.r.t various number of confounding contexts in Figure \ref{fig:a4}.
The experimental results indicate that the number of confounding contexts has consistent influence on the $NI$ of different classes.
Given any target class, we can simulate a more harsh distribution shift and further confuse the ConvNet by adding more confounding contexts.

\begin{figure}[htbp]
\centering 
\includegraphics[width=.64\linewidth]{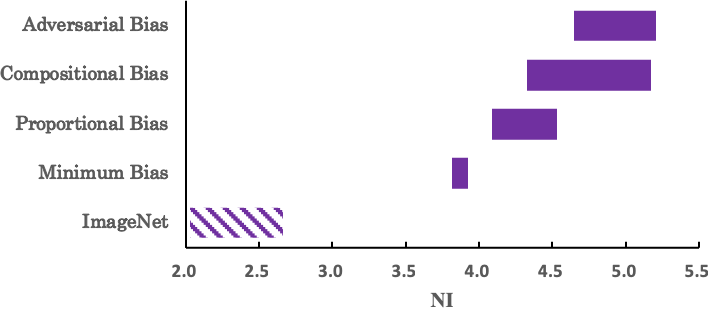}
\caption{Range of average $NI$ over $Animal$ superclass 
for different settings supported in NICO.}
\label{fig:cn}
\end{figure}

Finally, we show the range of NI in different Non-I.I.D. settings in Figure \ref{fig:cn}. 
We can see the level of NI in NICO is significantly higher than ImageNet, and there is an obvious ascending trend from Minimum Bias to Adversarial Bias settings.

% \subsection{Related Image Dataset}

% Although many acknowledged image datasets have been published, they cannot support the study of Non-I.I.D in terms of specified tasks  and properties.
% PASCAL VOC dataset, targeted at classification, detection and segmentation, have various tags for the tasks.
% But all the tags correspond to only 20 categories and images are well iconic, lack of rich and varied space to induce the Non-I.I.D .
% ImageNet, having promoted great advance in image classification, contains over 1 million images with labels and bounding boxes for 1000 classes.
% This super-large scale dataset still could not build the Non-I.I.D consciously, because nothing represents some kind of selection bias like context in NICO.
% MSCOCO is another large scale dataset which has over 300k images for 91 classes and more detailed tags for specified tasks, like human keypoints.
% Contexts, even captions, are supplied for recognition and caption tasks, but there are multiple vague semantic tags in one image, rather than unique and accurate ones in NCIO.
% None of these datasets can create kinds of controllable cases consciously and offer interpretable experimental results of Non-I.I.D like NICO.

\section{General Non-I.I.D. Image Classification}
In this section, we propose a novel model for General Non-I.I.D. image classification.

In the literature of Non-I.I.D. image classification, most previous methods are proposed for Targeted Non-I.I.D. image classification.
Domain adaptation and covariate shift methods \cite {long2017deep, long2015learning, sangineto2014we, tzeng2015simultaneous} are proposed to match distributions, transform feature space or learn invariant features  between training data and testing data.
These methods can achieve good performances but are less feasible in practice due to the fact that they need prior knowledge on testing data distribution.
On the other hand, several methods are proposed to liberalize the need of testing data information in Targeted Non-I.I.D. image classification.
For example, domain generalization methods \cite {ghifary2015domain, muandet2013domain} only use training data to learn a domain-agnostic model or invariant representations.
However, these methods about transfer learning \cite {pan2010survey} require the training data has multiple domains and we know which domain each sample belongs to. 
Moreover, the performance of these methods is highly dependent on the diversity of training data.

\begin{figure}[t]
\centering 
\includegraphics[width=.64\linewidth]{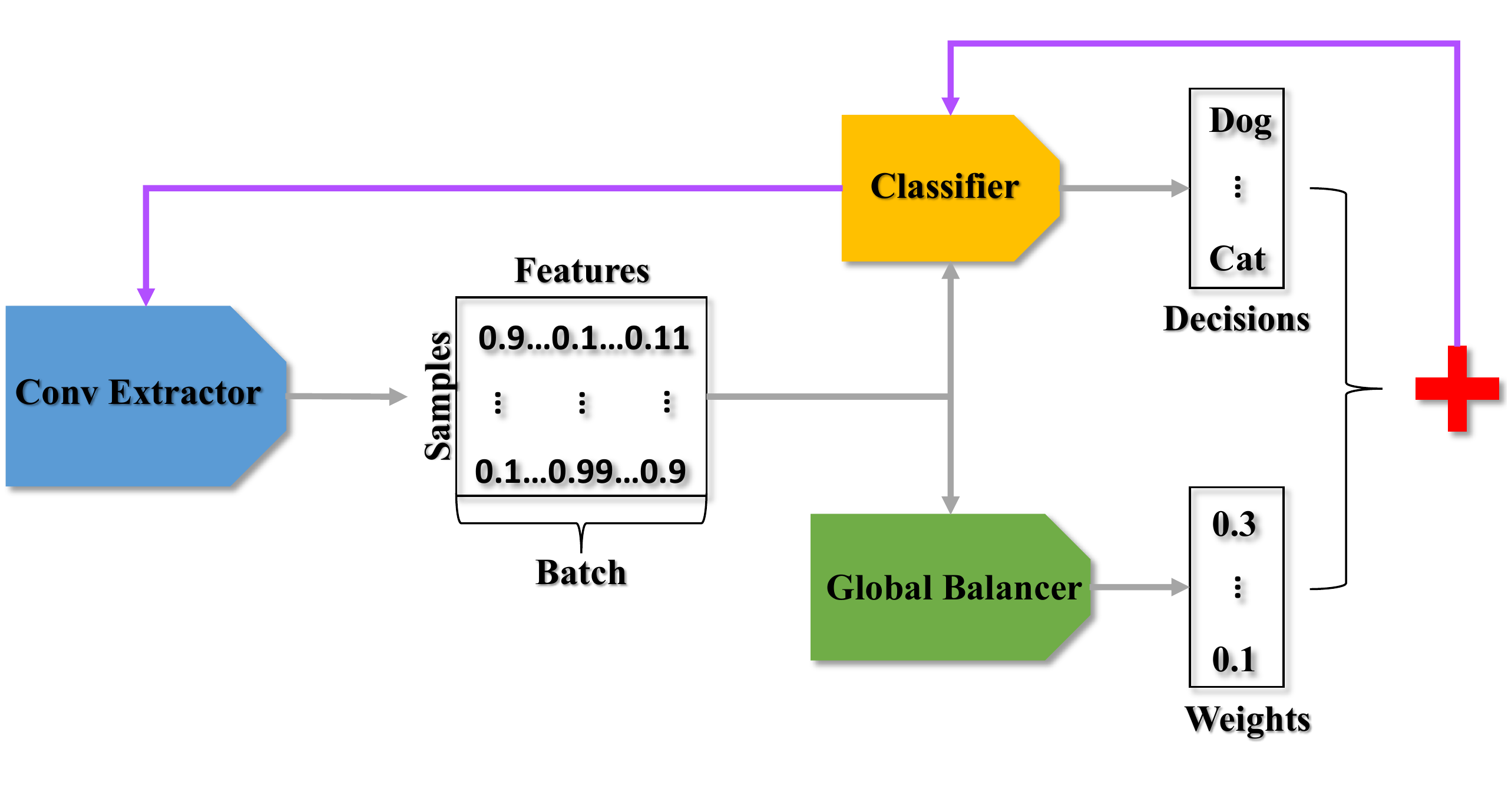}
\caption{Info flow in CNBB. The gray and purple lines refer to the forward and backward processes respectively.}
\label{fig:model}
\end{figure}

Recently, growing attention has been paid on General Non-I.I.D. learning.
In the literature of causality \cite{pearl2000causality},
an ideal model to resolve selection bias is to make policy based on causal variables, which keep stable across different domains\cite{rosenbaum1983central}. Popular methods based one observational data to estimate the causal effect of a treatment on the outcome include propensity score matching \cite{bang2005doubly, austin2011introduction}, markov blankets \cite{tsamardinos2003time, pellet2008using} and confounder balancing \cite{hainmueller2012entropy, kuang2017estimating} and etc \cite {li2018balancing}. Lately \cite {kuang2018stable} leverage causality for predictive modeling.
By performing global confounder balancing, one can accurately identify the stable features that are insensitive to unknown distribution shift for prediction.
\cite {shen2018causally} proposes a causally regularized logistic regression
called CRLR\footnote{\textbf{CRLR}: Causally Regularize Logistic Regression}for General Non-I.I.D. image classification
and achieve good performance in a relatively small dataset.
However, due to the lack of well-structured and reasonable-scaled dataset, these methods cannot leverage the powerful representation learning techniques (e.g. ConvNets) and therefore are not favourable for large-scale image classification tasks.

In this work, with the help of NICO, we extend the notion of global confounder balancing into ConvNet, and propose a novel model called CNBB, ConvNet with Batch Balancing.

\begin{algorithm}[htbp]
\caption{ConvNets with Batch Balancing (CNBB)}
   \label{alg:example}
\begin{algorithmic}
   \STATE {\bfseries Input:} Train dataset $D_{train}=\{(x_i, y_i)|i=1,...,n\}$
   \STATE {\bfseries Output:} Non-linear parameters $\theta$ and $\varphi$
   \STATE Initialize $\theta^{(0)}$, $\varphi^{(0)}$ and $t_1 \gets 0$
   \REPEAT
   \STATE Sample batch of images $\{(x_1, y_1), ..., (x_m, y_m)\}$
   \STATE Extract image features $\{g_{\varphi^{(t_1)}}(x_i), ..., g_{\varphi^{(t_1)}}(x_m)\}$
   \STATE Calculate indicator matrix $I$ of features
   \STATE Initialize sample weights $W^{(0)}$ and  $t_2 \gets 0$
   \REPEAT
   \STATE Optimize $W^{(t_2+1)}$ to minimize $Lossb$ in Eq.\ref{equationb}
   \STATE $t_2 \gets t_2+1$
   \UNTIL $Lossb$ converges or $t_2$ reaches maximum 
   \STATE Predict  $\{f_{\theta^{(t_1)}}(g_{\varphi^{(t_1)}}(x_1)), .., f_{\theta^{(t_1)}}(g_{\varphi^{(t_1)}}(x_m))\}$
   \STATE Optimize $\theta^{(t_1+1)}$ and $\varphi^{(t_1+1)}$ to minimize $Lossp$ in Eq.\ref{equationp}
   \STATE  $t_1 \gets t_1+1$
   \UNTIL $Lossp$ converges or $t_1$ reaches maximum
   \STATE {\bfseries return:} $\theta$ and $\varphi$
\end{algorithmic}
\end{algorithm}
 
\subsection{ConvNet with Batch Balancing}

The key idea in CRLR is global confounder balancing, which successively sets each feature as treatment variable, and learns an optimal set of sample weights that can balance the distribution of treated and control groups for any treatment variable.
Thereafter, the correlations among features will be disentangled and their true effects on class label can be more accurately estimated.

To introduce the notion of global confounder balancing into deep learning, we mainly face two challenges:

\begin{itemize}
    \item Confounder balancing methods assume features to be in binary form, while we generally have continuous features in ConvNet.
    \item For global confounder balancing, we need to learn a new set of sample weights for all the training samples in one iteration.
\end{itemize}

This is not feasible for ConvNet where we cannot feed all the training data into the model at once.

To overcome these challenges, we introduce a quantization loss for feature binarization and propose a batch confounder balancing method.
Specifically, given a batch of training images, we define the quantization loss as follows:
\begin{small}
\begin{equation}
\label{equationq}
Lossq=-\sum_{i=1}^{n}\left\|g_\varphi(x_i))\right\|_2^2,
\end{equation}
\end{small}
where n refers to the batch size, $x_i$ refers to the $i^{th}$ sample in a batch and $g_\varphi$ refers to the feature extractor (here we use the last FC layers in ConvNet as $g_\varphi$).
By minimizing $Lossq$, we can amplify the feature activated by tanh function from $(-1, 1)$ to approach to $\{-1, 1\}$.

Following the CRLR, we successively regard each feature as treatment, calculate the balancing loss of confounders and sum it over all the features globally.
Formally, we solve the batch confounder balancing problem as follows:
\begin{small}
\begin{equation}
\label{equationb}
\begin{aligned}
\begin{split}
\min_W Lossb=&\sum_{j=1}^{p}\left\|\frac{g_\varphi(X)_{-j}^{T}\cdot(W\odot I_j)}{W^{T}\cdot I_j}-\frac{g_\varphi(X)_{-j}^{T}\cdot(W\odot (1-I_j))}{W^{T}\cdot (1-I_j)}\right\|_2^2\\
&+\alpha\left\|W\right\|_2^2\ \ \ \ \ \ \ \ \ \ \ \ s.t.\ \sum_{i=1}^{n}W_i=1,\ W\geq0,
\end{split}
\end{aligned}
\end{equation}
\end{small}
where $W$ represents sample weights, $I_j$ means the $j^{th}$ column of $I$, and $I_{ij}$ refers to the treatment status of sample $i$ when setting feature $j$ as treatment variable, and $\left\|W\right\|_2^2$ can reduce the variance of weights to prevent the weights from overfitting outlier samples.
Different from CRLR, we define the confounder balancing loss w.r.t. a batch of training samples instead of the whole training samples.
Moreover, the sample weights and model parameters are jointly optimized through a supervised way in CRLR, while in CNBB we first fix the model parameters (a.k.a. representation) and learn the sample weights $W$ through an unsupervised way.

As far as we have learnt an optimal set of sample weights for a batch which can balance the confounder distribution, then we combine the weighted softmax loss and quantization loss and propose our CNBB model:
% \begin{small}
% \begin{equation}
% \label{equationc}
% Lossc=\sum_{i=1}^{n}w_i\ln(f_\theta(g_\varphi(x_i))\cdot y_i).
% \end{equation}
% \end{small}

\begin{small}
\begin{equation}
\label{equationp}
\min_{\theta, \varphi} Lossp =\sum_{i=1}^{n}w_i\ln(f_\theta(g_\varphi(x_i))\cdot y_i)+\lambda Lossq,
\end{equation}
\end{small}
where $f_\theta$ refers to softmax layer and $\lambda$ is a trade-off parameter between classification and quantization.

Algorithm \ref{alg:example} gives the complete steps of the batch balancing method and Figure \ref{fig:model} illustrates it intuitively.  

\subsection{Experiments on NICO}
In this section, we evaluate the proposed ConvNet with batch balancing (CNBB) in the task of General Non-I.I.D. image classification based on NICO.

\subsubsection{Experimental Settings}    

For fair comparison, we choose a typical structure of CNN and CNN with batch normalization \cite{ioffe2015batch} (CNN+BN) as baselines. 
The latter is a popular method in deep learning to improve the generalization ability of CNN by normalizing the scale of activations.
All the methods are implemented using PyTorch \cite{paszke2017automatic} and optimized by stochastic gradient descent.

We design four experiments according to the supported Non-I.I.D. settings of NICO in Sec 3.3:
\begin{itemize}
    \item Minimum bias (Exp 1): In this experiment, we randomly sample 8000 images for training and 2000 images for testing.
    \item Proportional bias (Exp 2): In this experiment, we fix the dominant ratio of training data to 5:1, and vary the dominant ratio of testing data from 1:5 to 4:1.  
    \item Compositional bias (Exp 3): In this experiment, we vary the number of contexts observed in training data from 3 to 7 while let all the contexts appear in testing data.
    \item Combined Proportional \& Compositional bias (Exp 4): To simulate a more harsh condition, for each class, we randomly select 7 contexts for training and the other 3 contexts for testing. Furthermore, we vary the dominant ratio of training data from 1:1 to 5:1 while fix the dominant ratio of testing data to 1:1.
\end{itemize}

% Experiment1 is carried out on animal superclass, the other two on vehicle superclass respectively.
% Furthermore, we can provide an insight of batch balancing by parameter sensitivity analysis.

%\begin{table}
%\begin{center}
%\begin{tabular}{|l|c|}
%\hline
%Exp1 & Testing Accuracy \\
%\hline\hline
%CNN &  \\
%CNN+BN &  \\
%CNBB &      \\
%\hline
%\end{tabular}
%\end{center}
%\caption{Performances of different methods %on test accuracy (\%)  for animal in %proportion bias setting for animal.}
%\label{table3}
%\end{table}

% \begin{table}
% \begin{center}
% \begin{tabular}{|l|c|}
% \hline
% Exp1 & Testing Accuracy \\
% \hline\hline
% CNN & 49.60\% \\
% CNN+BN & 46.48\% \\
% CNBB & \textbf{49.94\%} \\
% \hline
% \end{tabular}
% \end{center}
% \caption{Performances of different methods on test accuracy (\%) for minimun bias in $Animal$ superclass.}
% \label{table7}
% \end{table}

\begin{table}[htbp]
\begin{center}
\begin{tabular}{|l|c|c|c|c|c|}
\hline
Exp2 & 1 : 5 & 1 : 1 & 2 : 1 & 3 : 1 & 4 : 1 \\
\hline\hline
CNN & 37.17 & 37.80 & 41.46 & 42.50 & 43.23 \\
CNN+BN & 38.70 & \textbf{39.60} & 41.64 & 42.00 & 43.85 \\
CNBB & \textbf{39.06} & \textbf{39.60} & \textbf{42.12} & \textbf{43.33} & \textbf{44.15} \\
\hline
\end{tabular}
\end{center}
\caption{Performances of different methods on test accuracy (\%) for proportional bias in $Animal$ superclass.}
\label{table3}
\end{table}

\begin{table}[htbp]
\begin{center}
\begin{tabular}{|l|c|c|c|c|c|}
\hline
Exp3 & 3 & 4 & 5 & 6 & 7 \\
\hline\hline
CNN & 40.61 & 42.32 & 43.34 & 44.03 & 44.03 \\
CNN+BN & \textbf{41.98} & 38.85 & 43.12 & 44.71 & 44.31 \\
CNBB & 41.41 & \textbf{43.34} & \textbf{44.54} & \textbf{45.96} & \textbf{45.16} \\
\hline
\end{tabular}
\end{center}
\caption{Performances of different methods on test accuracy (\%) for composional bias in $Vehicle$ superclass.}
\label{table4}
\end{table}
\begin{table}[htbp]
\begin{center}
\begin{tabular}{|l||c|c|c|c|c|}
\hline
Exp4 & 1 : 1 & 2 : 1 & 3 : 1 & 4 : 1 & 5 : 1 \\
\hline\hline
CNN & 37.07 & 35.20 & 34.53 & 34.13 & 33.73 \\
CNN+BN & 33.87 & 32.93 & 31.20 & 30.93 & 30.67 \\
CNBB & \textbf{38.98} & \textbf{36.89} & \textbf{35.87} & \textbf{35.33} & \textbf{35.02} \\
\hline
\end{tabular}
\end{center}
\caption{Performances of different methods of test accuracy (\%) for combined proportional \& compositional bias in $Vehicle$ superclass.} 
\label{table5} 
\end{table}

\subsubsection{Experimental Results}
We calculate the average testing accuracy of all the methods for each experiment.
First of all, CNBB is comparable with CNN in the minimum bias setting, with a slightly higher accuracy ($49.94\%$ v.s. $49.60\%$), and CNN+BN performs worst ($46.48\%$).
For the other three experiments with explicit distribution shift between training data and testing data, CNBB outperforms the other baselines at almost every setting, as shown in Table \ref{table3},\ref{table4},\ref{table5}, indicating its effectiveness in Non-I.I.D. image classification.
Note that the performance of CNN with batch normalization is relatively unstable compared to original CNN across different experiments.
It is mainly because, in the General Non-I.I.D. setting, the agnostic distribution shift between training and testing data cannot be effectively normalized only based on the training data.
Comparatively, the batch balancing module enable CNBB to identify more stable features and therefore resist the negative effect brought by distribution shift to some extent.

\begin{table} [htbp]
\label{table6}
\begin{center} 
\setlength{\tabcolsep}{6mm}{
\begin{tabular}{|c|c|c|}
\hline
Experiment & Improvement & $NI$ \\
\hline 
Exp1 & 0.33\% & 3.81 - 3.93 \\
\hline
Exp2 & 1.22\% & 4.17 - 4.53 \\
\hline
Exp3 & 1.22\% & 4.13 - 4.34 \\
\hline
Exp4 & 1.49\% & 4.44 - 4.90 \\
\hline
\end{tabular}}
\end{center}
\caption{The range of NI with respect to the average improvement of performance to CNN.}
\label{table6}
\end{table}

We further summarize the improvement of CNBB over the best baseline in different experiments.
From Table \ref{table6}, we can clearly find that with the discrepancy between the training and testing data getting larger (indicated by higher $NI$), CNBB gains larger improvement over baselines, which demonstrate the advantage of our method in more challenging Non-I.I.D. settings.

\begin{figure}
\centering 
\includegraphics[width=.64\linewidth]{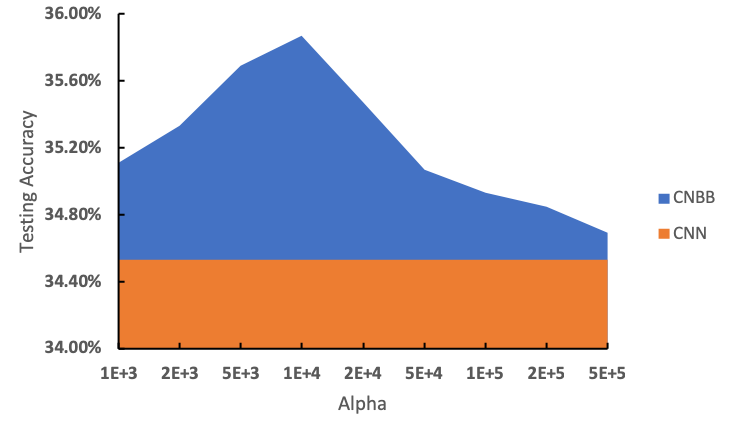}
\caption{Parameter sensitivity analysis of Exp4. Testing accuracy with respect to the trade-off parameter $\lambda$ in Eq.\ref{equationb} while we set dominant ratio of training data to 3:1. The blue area represents the improvement of CNBB against CNN.}
\label{fig:pa}
\end{figure}

Finally, we analyze the hyperparameter $\alpha$. $\alpha$ eventually plays the role of trading-off the valid sample size and degree of batch balancing. In theory, when $\alpha$ is extremely large, the weights of samples tend to be uniform, resulting in a largest valid sample size. When $\alpha$ is zero, the algorithm tend to converge to a situation where sample weights concentrate on only a few images, but lead to an optimal batch balancing.
Both of valid sample size and degree of batch balancing are critical for the performances of Non-I.I.D. image classification.
As in Eq \ref {equationb}, we tune the hyperparameter $\alpha$ with 9 values (1e3 to 5e5) in all the experiments.
Taking the case where training dominant ratio is 3:1 in Table \ref {table5} as an example, a convex hull is clear in Figure \ref {fig:pa}. 
Along with the increasing $\alpha$, the gain of CNBB will tend to vanish eventually. The results fully demonstrate the effectiveness of batch balancing module.

\section{Conclusion and Future Works}
In this paper, we introduce a new dataset NICO for promoting the research on Non-I.I.D. image classification.
To the best of our knowledge, NICO is the first well-structured Non-I.I.D. image dataset with reasonable scale to support the training of ConvNets.
By incorporating the idea of context, NICO can provide various Non-I.I.D. settings and create different levels of Non-IIDness consciously.
We also propose a simple baseline model with ConvNet structure for General Non-I.I.D. image classification problem, where testing data bear agnostic distribution shift from training data.
Empirical results clearly demonstrate the capability of NICO on training the ConvNets and the superiority of the proposed model in various Non-I.I.D. settings.

Our future works will focus on the followings. 
Firstly, both quality and quantity of NICO continue to be improved. 
Orthogonal contexts, denoised images and proper area ratio of objects will be explored to make NICO more controllable to tune bias and response to the Non-I.I.D uniquely.
And we will expand the scale of dataset from all the levels for adequate demands. 
Secondly, more settings about different forms of Non-I.I.D are expected to be exploited.
So other visual concepts may be added to NICO if needed and the ways of using NICO to meet new settings will be given in detail. Thirdly, more effective models will be designed for addressing problems in different settings of Non-I.I.D image classification.

\clearpage
\setcounter {table} {0} 
\section{Appendix}
\makeatletter\def\@captype{table}\makeatother
\caption {Basic structure of ConvNet used in this paper.}
\begin {center}
\vspace{10pt}
\begin {tabular}{|c|c|c|c|}
\hline 
\multicolumn{3}{|c|}{Structure of ConvNet} \\
\hline 
Layer & Filter & height \& width \\
\hline
input   & 3    & (64 * 64)    \\
\hline
conv    & 64   & (64 * 64)    \\
\hline
\multicolumn{3}{|c|}{relu}    \\
\hline
maxpool & 64   & (32 * 32)    \\
\hline
conv    & 128  & (32 * 32)    \\
\hline
\multicolumn{3}{|c|}{relu}    \\
\hline
maxpool & 128  & (16 * 16)    \\
\hline
conv    & 256  & (16 * 16)    \\
\hline
\multicolumn{3}{|c|}{relu}    \\
\hline
maxpool & 256  & (8  * 8 )    \\
\hline
conv    & 512  & (8  * 8 )    \\
\hline
\multicolumn{3}{|c|}{relu}    \\
\hline
maxpool & 512  & (4  * 4 )    \\
\hline
conv    & 1024 & (4  * 4 )    \\
\hline
\multicolumn{3}{|c|}{relu}    \\
\hline
maxpool & 1024 & (2  * 2 )    \\
\hline
fc      & 512  & 1            \\
\hline
\multicolumn{3}{|c|}{relu}    \\
\hline
fc      & 50   & 1            \\
\hline
\multicolumn{3}{|c|}{tanh}    \\
\hline
fc      & 10/9 & 1            \\
\hline
\multicolumn{3}{|c|}{softmax} \\
\hline
\end {tabular}
\end {center}

\begin {sidewaystable}
\caption {Data size of each context for every class in $Animal$ superclass.}
\begin {center}
\begin{scriptsize}
\begin{sc}
\begin {tabular}{|l|c|c|c|c|c|c|c|c|c|c|}
\hline
\multicolumn{11}{|c|}{$Animal$} \\
\hline
\multirow {2}{*}{Bear} & black & brown & eating grass & in forest & in water & lying & on ground & on snow & on tree & white \\ 
\cline {2-11}
& 245 & 220 & 133 & 243 & 169 & 217 & 97 & 111 & 70 & 104 \\
\hline
\multirow {2}{*}{Bird} & eating & flying & in cage & in hand & in water & on branch & on grass & on ground & on shoulder & standing \\ 
\cline {2-11}
& 187 & 203 & 90 & 94 & 81 & 239 & 242 & 276 & 77 & 101 \\
\hline
\multirow {2}{*}{Cat} & at home & eating & in cage & in river & in street & in water & on grass & on snow & on tree & walking \\ 
\cline {2-11}
& 274 & 270 & 109 & 141 & 177 & 50 & 140 & 137 & 50 & 131 \\
\hline
\multirow {2}{*}{Cow} & aside people & at home & eating & in forest & in river & lying & on grass & on snow & spotter & standing \\ 
\cline {2-11}
& 56 & 77 & 147 & 131 & 139 & 162 & 147 & 135 & 75 & 123 \\
\hline
\multirow {2}{*}{Dog} & at home & eating & in cage & in street & in water & lying & on beach & on grass & on snow & running \\ 
\cline {2-11}
& 92 & 264 & 122 & 87 & 139 & 143 & 280 & 158 & 238 & 101 \\
\hline
\multirow {2}{*}{Elephant} & eating & in circus & in forest & in river & in street & in zoo & lying & on grass & on snow & standing \\ 
\cline {2-11}
& 122 & 114 & 160 & 178 & 90 & 162 & 69 & 103 & 69 & 111 \\
\hline
\multirow {2}{*}{Horse} & aside people & at home & in forest & in river & in street & lying & on beach & on grass & on snow & running \\ 
\cline {2-11}
& 124 & 86 & 146 & 73 & 77 & 141 & 165 & 165 & 138 & 143 \\
\hline
\multirow {2}{*}{Monkey} & climbing & eating & in cage & in forest & in water & on beach & on grass & on snow & sitting & walking \\ 
\cline {2-11}
& 88 & 168 & 77 & 140 & 118 & 50 & 106 & 102 & 168 & 100 \\
\hline
\multirow {2}{*}{Rat} & at home & eating & in cage & in forest & in hole & in water & lying & on grass & on snow & running \\ 
\cline {2-11}
& 126 & 169 & 57 & 85 & 50 & 85 & 50 & 124 & 50 & 50 \\
\hline
\multirow {2}{*}{Sheep} & aside people & at sunset & eating & in forest & in water & lying & on grass & on road & on snow & walking \\ 
\cline {2-11}
& 50 & 66 & 116 & 95 & 71 & 109 & 132 & 111 & 87 & 81 \\
\hline
\end {tabular}
\end{sc}
\end{scriptsize}
\end {center}
\end {sidewaystable}

\begin {sidewaystable}
\caption {Data size of each context for every class in $Vehicle$ superclass.}
\begin {center}
\begin{scriptsize}
\begin{sc}
\begin {tabular}{|l|c|c|c|c|c|c|c|c|c|c|}
\hline
\multicolumn{11}{|c|}{$Vehicle$} \\
\hline
\multirow {2}{*}{Airplane} & around cloud & aside
mountain & at airport & at night & in city & in sunrise & on beach & on grass & taking off & with pilot \\ 
\cline {2-11}
& 87 & 76 & 153 & 76 & 55 & 70 & 104 & 53 & 128 & 128 \\
\hline
\multirow {2}{*}{Bicycle} & in garage & in street & in sunset & on beach & on grass & on road & on snow & shared & velodrome & with people \\ 
\cline {2-11}
& 143 & 113 & 134 & 131 & 219 & 125 & 163 & 225 & 220 & 166 \\
\hline
\multirow {2}{*}{Boat} & at wharf & cross bridge &  in city & in river & in sunset & on beach & sailboat & with people & wooden & yacht \\ 
\cline {2-11}
& 219 & 190 & 194 & 265 & 196 & 168 & 252 & 143 & 248 & 281 \\
\hline
\multirow {2}{*}{Bus} & aside traffic light & aside tree & at station  & at yard & double decker & in city & on bridge & on snow & with people & \\ 
\cline {2-11}
& 35 & 165 & 95 & 74 & 221 & 199 & 45 & 124 & 51 & \\
\hline
\multirow {2}{*}{Car} & at park & in city & in sunset & on beach & on booth & on bridge & on road & on snow & on track & with people\\ 
\cline {2-11}
& 80 & 149 & 89 & 102 & 112 & 36 & 146 & 184 & 89 & 39 \\
\hline
\multirow {2}{*}{Helicopter} & aside mountain & at heliport & in city & in forest & in sunset & on beach & on grass & on sea & on snow & with people\\ 
\cline {2-11}
& 165 & 185 & 69 & 124 & 160 & 107 & 147 & 156 & 180 & 58 \\
\hline
\multirow {2}{*}{Motorcycle} & in city & in garage & in street & in sunset & on beach & on grass & on road & on snow & on track & with people \\ 
\cline {2-11}
& 194 & 148 & 173 & 157 & 122 & 99 & 162 & 134 & 185 & 168 \\
\hline
\multirow {2}{*}{Train} & aside mountain & at station & cross tunnel & in forest & in sunset & on beach & on bridge & on snow & subway & \\ 
\cline {2-11}
& 63 & 158 & 36 & 100 & 94 & 46 & 54 & 129 & 70 &  \\
\hline
\multirow {2}{*}{Truck} & aside mountain & in city & in forest & in race & in sunset & on beach & on bridge & on grass & on road & on snow \\ 
\cline {2-11}
& 62 & 77 & 91 & 134 & 155 & 97 & 44 & 78 & 145 & 117 \\
\hline
\end {tabular}
\end{sc}
\end{scriptsize}
\end {center}
\end {sidewaystable}

\section*{References}
\bibliographystyle{elsarticle-num}
\bibliography{egbib}

\end {document}